\documentclass[letterpaper, 10 pt, conference]{ieeeconf}  

\IEEEoverridecommandlockouts                              

\usepackage{amsmath} 
\usepackage{amssymb}  
\usepackage[T1]{fontenc}
\usepackage{multirow}
\usepackage{graphicx} 
\usepackage{url}
\usepackage{bm}
\usepackage{cite}
\usepackage[hidelinks]{hyperref}

\title{\LARGE \bf
Open-RGBT: Open-vocabulary RGB-T Zero-shot Semantic Segmentation in Open-world Environments}

\author{Meng Yu,  Luojie Yang, Xunjie He, Yi Yang, Yufeng Yue*, ~\IEEEmembership{Member,~IEEE}
\thanks{This work is supported by the National Natural Science Foundation of China under Grant 92370203, 62233002. 
(Corresponding Author: Yufeng Yue, yueyufeng@bit.edu.cn)}
\thanks{Meng Yu, Luojie Yang, Xunjie He, Yi Yang, and Yufeng Yue are with School of Automation, Beijing Institute of Technology, Beijing 100081, China. }%
}

\begin{document}

\maketitle
\thispagestyle{empty}
\pagestyle{empty}

\begin{abstract}

Semantic segmentation is a critical technique for effective scene understanding. Traditional RGB-T semantic segmentation models often struggle to generalize across diverse scenarios due to their reliance on pretrained models and predefined categories. Recent advancements in Visual Language Models (VLMs) have facilitated a shift from closed-set to open-vocabulary semantic segmentation methods. However, these models face challenges in dealing with intricate scenes, primarily due to the heterogeneity between RGB and thermal modalities. To address this gap, we present Open-RGBT, a novel open-vocabulary RGB-T semantic segmentation model. Specifically, we obtain instance-level detection proposals by incorporating visual prompts to enhance category understanding. Additionally, we employ the CLIP model to assess image-text similarity, which helps correct semantic consistency and mitigates ambiguities in category identification. Empirical evaluations demonstrate that Open-RGBT achieves superior performance in diverse and challenging real-world scenarios, even in the wild, significantly advancing the field of RGB-T semantic segmentation. The project page of Open-RGBT is available at \href{https://OpenRGBT.github.io/}{https://OpenRGBT.github.io/}.

\end{abstract}

\section{INTRODUCTION}

Semantic segmentation plays a fundamental role in enabling scene understanding for unmanned systems, especially in practical applications such as autonomous driving \cite{book}, robotic manipulation \cite{robot}, and remote sensing \cite{remote}. Although previous works \cite{segnet, deeplab, sam} have achieved remarkable segmentation performance on standard RGB-based datasets, they often falter under conditions of reduced visibility caused by adverse weather \cite{vifnet} or in poor illumination settings \cite{sambad}. This challenge is compounded by the need for open-world adaptability \cite{yueicra}. To address these limitations, many researchers have introduced thermal/infrared images to enhance the performance of visual perception tasks \cite{mfnet, rtfnet, pst900, gmnet, xunjiergbt}. Despite significant progress, these semantic segmentation models are predominantly trained on pre-defined categories, limiting their ability to generalize to unseen classes.

\begin{figure}[!ht]
  \centering
  \setlength{\abovecaptionskip}{-0.2em}
  \includegraphics[width=\linewidth]{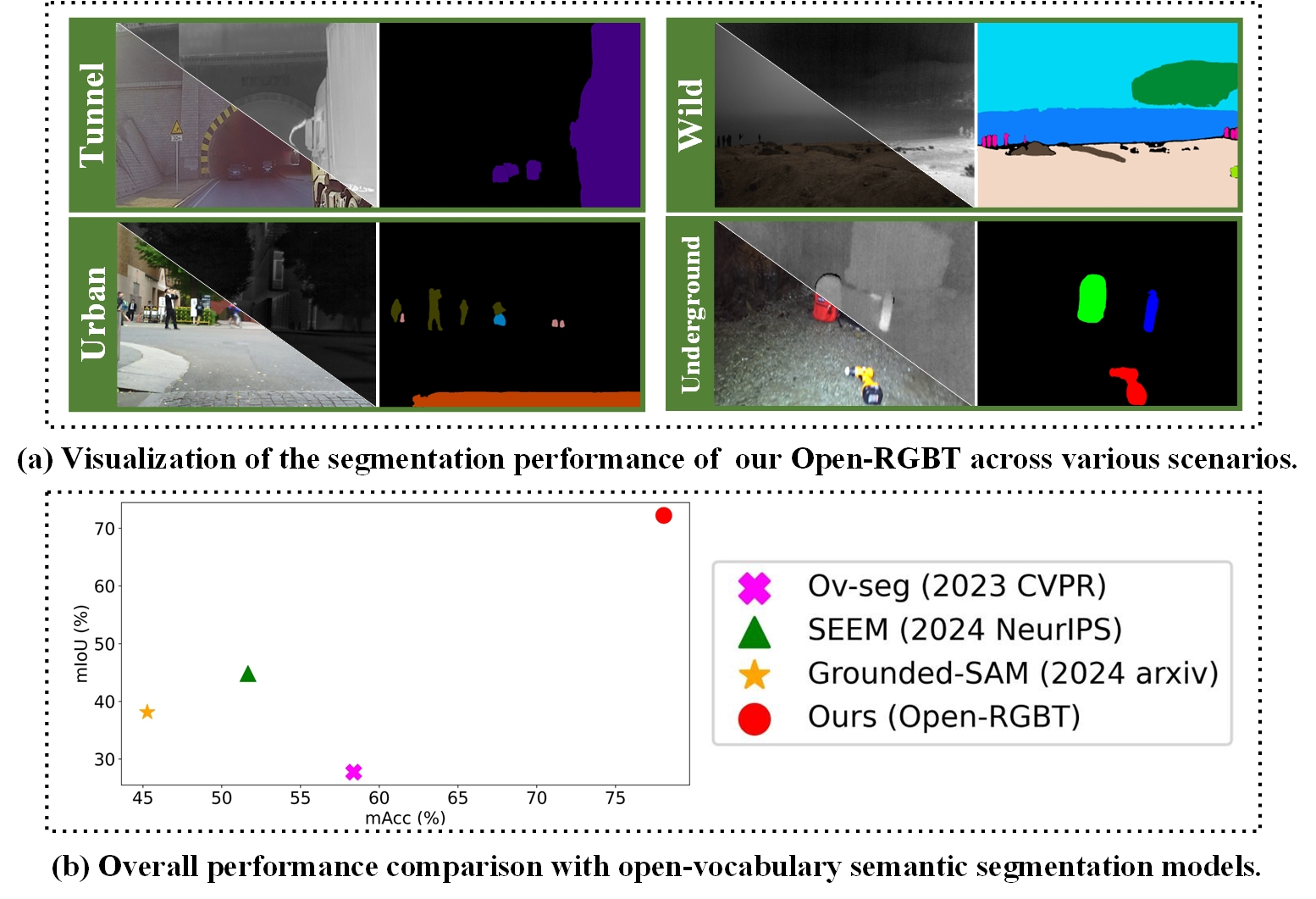}
  \caption{We introduce \textbf{Open-RGBT}, a novel open-vocabulary RGB-T semantic segmentation model, which facilitates zero-shot semantic segmentation across various open-world scenarios.}
  \vspace{-1.5em}
  \label{fig1}
\end{figure}

Open-vocabulary learning \cite{ovlearning} has been proposed to address the aforementioned issues and to extend the capabilities of segmentation tasks. With the rapid development of pre-trained Visual Language Models (VLMs) such as CLIP \cite{clip}, previous works \cite{lseg, openseg, zsseg, zegformer, ovseg} have leveraged these models' open-vocabulary classification capabilities to implement RGB image segmentation. The VLMs provide a new perspective for achieving RGB-T zero-shot semantic segmentation, but extending these models to the RGB-T domain still faces challenges. For example, in these approaches, ZSSeg\cite{zsseg}, ZegFormer \cite{zegformer}, and Ov-seg \cite{ovseg} assign semantics to the mask proposals by applying pre-trained CLIP model. However, the effectiveness of these models relies heavily on the accurate generation and classification of masks. Since CLIP is trained exclusively on RGB images, the domain gap between detected proposals and the training datasets often leads to suboptimal classification of masked images. On the other hand, Grounded-SAM \cite{groundsam} employs the open-vocabulary detector Grounding DINO \cite{groundingdino} to generate detection proposals, which are then refined using SAM \cite{sam} for segmentation. However, the performance of Grounded-SAM also hinges on the accurate detection of objects and may struggle with ambiguous category understanding due to the heterogeneity among different modalities. 

These challenges highlight that both misclassifications and missed detections can result in an incomplete understanding. Moreover, relying on a single modality is insufficient for models to comprehensively understand the open world. To address these limitations, we propose Open-RGBT, a zero-shot framework designed to facilitate semantic understanding of arbitrary text queries.

Specifically, we first generate instance-level detection proposals from the fused images using an off-the-shelf foundation model. However, this process may result in an ambiguous understanding of specific categories in RGB-T datasets, owing to variations in data distribution. To address this, we incorporate visual prompts to enhance the detection of novel or complex categories. Unlike previous methods that may miss detections, our approach allows for the acquisition of more coarse results even in such cases. Additionally, recognizing that these coarse detections may carry semantic ambiguity, we employ the CLIP model to ensure semantic consistency in the detection proposals. This strategy enables Open-RGBT to achieve accurate detection results, which can subsequently be segmented and captioned for a deeper understanding of the scene. Unlike dataset-specific models, Open-RGBT does not require extensive training and is adaptable to a wide array of datasets. To foster community engagement, we have assembled a collection of 273 paired RGB-T images in the real-world, along with corresponding semantic ground truth, called MSVID dataset. This dataset encompasses various road conditions, including rain, haze, and varying illumination.

In summary, the contributions of this paper are summarized as follows:
\begin{enumerate}
    \item We introduce open-vocabulary models for RGB-T semantic segmentation, enhancing the generalization capabilities and semantic segmentation performance. 
    \item The model incorporates optimized visual prompts and semantic consistency correction module, enabling a deeper understanding and precise categorization of novel or unique classes within specific datasets. 
    \item We release a multi-weather RGB-T dataset with semantic annotations in open environments. Extensive evaluations indicate that Open-RGBT achieves superior zero-shot performance in real-world scenes.
\end{enumerate}

\section{RELATED WORK}
\subsection{Closed-set RGB-T Semantic Segmentation}
Single-modal semantic segmentation networks, such as SegNet \cite{segnet} and DeepLab \cite{deeplab}, are often vulnerable to extreme conditions, like adverse weather or poor illumination conditions. To overcome these limitations, researchers have explored multimodal fusion techniques to exploit the complementary information. Specifically, they proposed various types of feature
fusion modules, such as the heterogeneous feature-level fusion \cite{mfnet, rtfnet, pst900}, multi-scale feature fusion \cite{LASnet, semanticrt, eaefnet, cainet}, and attention-weighted fusion \cite{gmnet, egfnet, ccffnet}.

Although the above methods have been implemented to obtain more complementary and rich information, leading to outstanding performance on MFNet dataset \cite{mfnet} and PST900 dataset \cite{pst900}, these models have poor generalization on new datasets with different data distribution. Meanwhile, they need to train separate models on different datasets, increasing the computational cost and complexity. Besides, these models are mainly trained with pre-defined categories, which are hard to generalize to unseen classes.

\subsection{Open-vocabulary Semantic Segmentation}
To overcome the limitation of previous closed-set segmentation methods, open-vocabulary semantic segmentation aims to understand an image with arbitrary categories described by texts. With the impressive progress of pre-trained VLMs (\textit{i}.\textit{e}., CLIP \cite{clip}, ALIGN\cite{align}), an increasing number of methods are attempting to extend their semantic understanding capabilities to vision perception tasks. LSeg \cite{lseg} and OpenSeg \cite{openseg} align the segment-level visual features with text embeddings via region-word grounding. 

Subsequent work (ZSSeg \cite{zsseg}, ZegFormer \cite{zegformer}, and Ov-seg \cite{ovseg}) decouple the problem into a two-stage task, namely class-agnostic segmentation and mask classification. They firstly generate class-agnostic mask proposals, and then utilize pre-trained CLIP model to obtain language-aligned visual features and classify each mask. In contrast, Grounded-SAM \cite{groundsam} uses Grounding DINO \cite{groundingdino} as an open-set object detector to generate detection boxes based on arbitrary text inputs, and then leverages SAM \cite{sam} model to segment these regions. Notably, these models are limited by accurate proposal generation. Furthermore, the inherent heterogeneity between modalities may hinder category understanding in the thermal domain. In an open-world context, it is imperative for the model to exhibit greater adaptability.

To address the above challenges, Open-RGBT takes the paired RGB-T images as input and leverages a two-stage pipeline. Instead, with joint visual prompts and semantic consistency correction, Open-RGBT facilitates open-vocabulary RGB-T semantic segmentation in complex scenes.

\section{METHODOLOGY}
\subsection{Framework Overview}
Open-RGBT takes a pair of visible-thermal images $\mathbf{I}=[\mathbf{I}_{\mathrm{RGB}}, \mathbf{I}_{\mathrm{T}}]$ with a predefined set of semantic classes $\mathbf{T}=\left \{t_1, t_2, \dots, t_K\right \}$ as input, and produces a set of $N$ mask proposals $\mathbf{M}_n$ and corresponding caption predictions $\mathbf{C}_n$ of the observed scene as output, where $n=1, 2, \dots, N$.

\begin{figure*}[!ht]
  \centering
  \includegraphics[width=\textwidth]{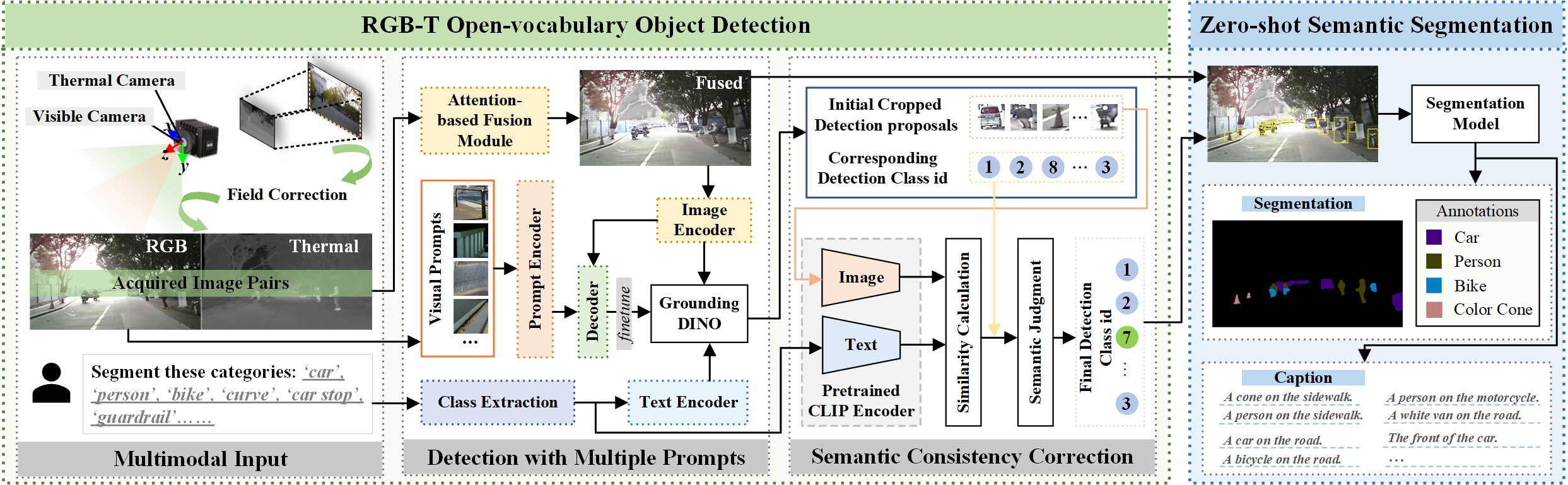}
  \caption{The overall framework of Open-RGBT consists of two stages: RGB-T Open-vocabulary Object Detection and Scene Semantic Understanding.}
  \vspace{-1.5em}
  \label{fig2}
\end{figure*}

The overall Open-RGBT framework, as depicted in Fig. \ref{fig2}, consists of two successive stages. The first one generates detection boxes for objects or regions from fused images by leveraging the textual information and visual prompts as conditions. Then, the detected proposals serve as the prompts for the segmentation model to generate precise mask annotations. By sequentially executing the two stages, Open-RGBT acquires a profound understanding of the scene.

\subsection{RGB-T Open-vocabulary Object Detection}
This stage takes paired RGB-T images as input and utilizes classes as text prompts to generate bounding boxes. It is composed of three parts: Attention-based RGB-T Adaptive Fusion, Detection with Multiple Prompts, and Semantic Consistency Correction Module (SCCM). Firstly, RGB-T images are adaptively fused, ensuring that the most informative features from each modality are preserved and integrated into a single, enhanced image representation. Secondly, visual and text prompts are incorporated to better understand and detect novel or complex categories. Lastly, in addition to Grounding DINO's prediction, we employ CLIP model to correct the semantic consistency of the detection proposals, so as to avoid semantic ambiguity and achieve accurate results.

\subsubsection{Attention-based RGB-T Adaptive Fusion}
Considering the varying richness and informativeness of different modalities present in the input images, we employ attention mechanism to obtain dynamic fusion weights for combining these two modalities. Inspired by PSFusion \cite{psf}, we adopt the image fusion path from the the scene restoration branch. Given a pair of registered visible image $\mathbf{I}_{\mathrm{RGB}} \in \mathbb{R}^{H \times W \times 3}$ and thermal image $\mathbf{I}_{\mathrm{T}} \in \mathbb{R}^{H \times W \times 1}$, the image fusion path ultimately synthesizes the fused image $\mathbf{I}_{f} \in \mathbb{R}^{H \times W \times 3}$. Here, the global-local attention mechanism is employed, focusing on both global context and local details based on the input data. By updating the fusion weights in this self-adaptive manner, the proposed fusion approach can maintain effective fusion even under conditions where the quality of one modality may be compromised or suboptimal.

\subsubsection{Detection with Multiple Prompts}
With the the fused image $\mathbf{I}_{f}$ and given categories $\mathbf{T}$, we can feed them together into the Grounding DINO \cite{groundingdino} model to obtain the coarse detection boxes and corresponding class id. Although Grounding DINO can yield excellent performance in open-set detection, it may have an ambiguous understanding of specific categories in certain datasets due to the disparity in data distribution, leading to false or missed detections. 

To address this issue, we incorporate visual prompts to enhance object understanding. When encountering a novel category or one whose semantics are inconsistent with predefined categories, only several visual examples are required to understand the category in an in-context manner. 

Specifically, the T-Rex2 \cite{trex} model is leveraged to generate visual prompt embeddings and infer the detection boxes. Firstly, by cropping the desired area, we can obtain $J$ specified 4$D$ normalized boxes $b_j=(x_j, y_j, w_j, h_j)$, where $j=1, 2, \dots, J$, $x_j$, $y_j$ represent the horizontal and vertical coordinates in the upper left corner of the bounding box, respectively, and $w_j$, $h_j$ represents the width and height of the bounding box, respectively. These coordinate inputs are encoded into position embeddings through a fixed sine-cosine embedding layer. Subsequently, a linear layer projectd these embeddings into a uniform dimension:
\begin{equation}
B=\mathrm{Linear}(\mathrm{PE}(b_1, \dots, b_J); \theta _B): \mathbb{R}^{J\times 4D}\rightarrow \mathbb{R}^{J\times D}
\label{eq1}
\end{equation}
where $\mathrm{PE}$ stands for position embedding and $\mathrm{Linear}(\cdot; \theta)$ indicate a linear project operation with parameter $\theta$. 

Next, to aggregate features from other visual prompts, the T-Rex2 model introduces global position embeddings $B'$, derived from global normalized coordinates [0.5, 0.5, 1, 1], thereby constructing the input query embedding $Q$. For the $j$-th prompt, the query feature $Q_j^{'}$ after cross attention is computed as:
\begin{equation}
Q_j^{'}=\mathrm{MSDeformAttn}(Q_j, b_j, \left \{ \bm{f}_l \right \}_{l=1}^{L})
\label{eq2}
\end{equation}
where $\bm{f}_l\in \mathbb{R}^{C_l \times H_l \times W_l}$ ($l=1, 2, \dots, L$) denotes the feature maps output from the image encoder, and $L$ is the number of feature map layers. 

Then, a self-attention layer to regulate the relationships among different queries and a feed-forward layer for projection are employed. The global content query output is used as the final visual prompt embedding $V$:
\begin{equation}
V=\mathrm{FFN}(\mathrm{SelfAttn(Q^{'})})[-1]
\label{eq3}
\end{equation}

Finally, a DETR-like decoder is employed for box prediction. The whole detection proposals can be expressed as:
\begin{equation}
v_N=\mathrm{Det_{vp}}(V, \mathbf{I}_{f})\cup \mathrm{Det_{gd}}(\mathbf{T}, \mathbf{I}_{f})
\label{eq4}
\end{equation}
where $N$ denotes the number of detection proposals, $\mathrm{Det_{vp}}(\cdot,\cdot)$ and $\mathrm{Det_{gd}}(\cdot,\cdot)$ represent the detection process with visual prompts and with Grounding DINO, respectively. By integrating text and visual prompts, the precision and accuracy of the detection model can be significantly improved.

\subsubsection{Semantic Consistency Correction}
\begin{figure}[!ht]
  \centering
  \includegraphics[width=\linewidth]{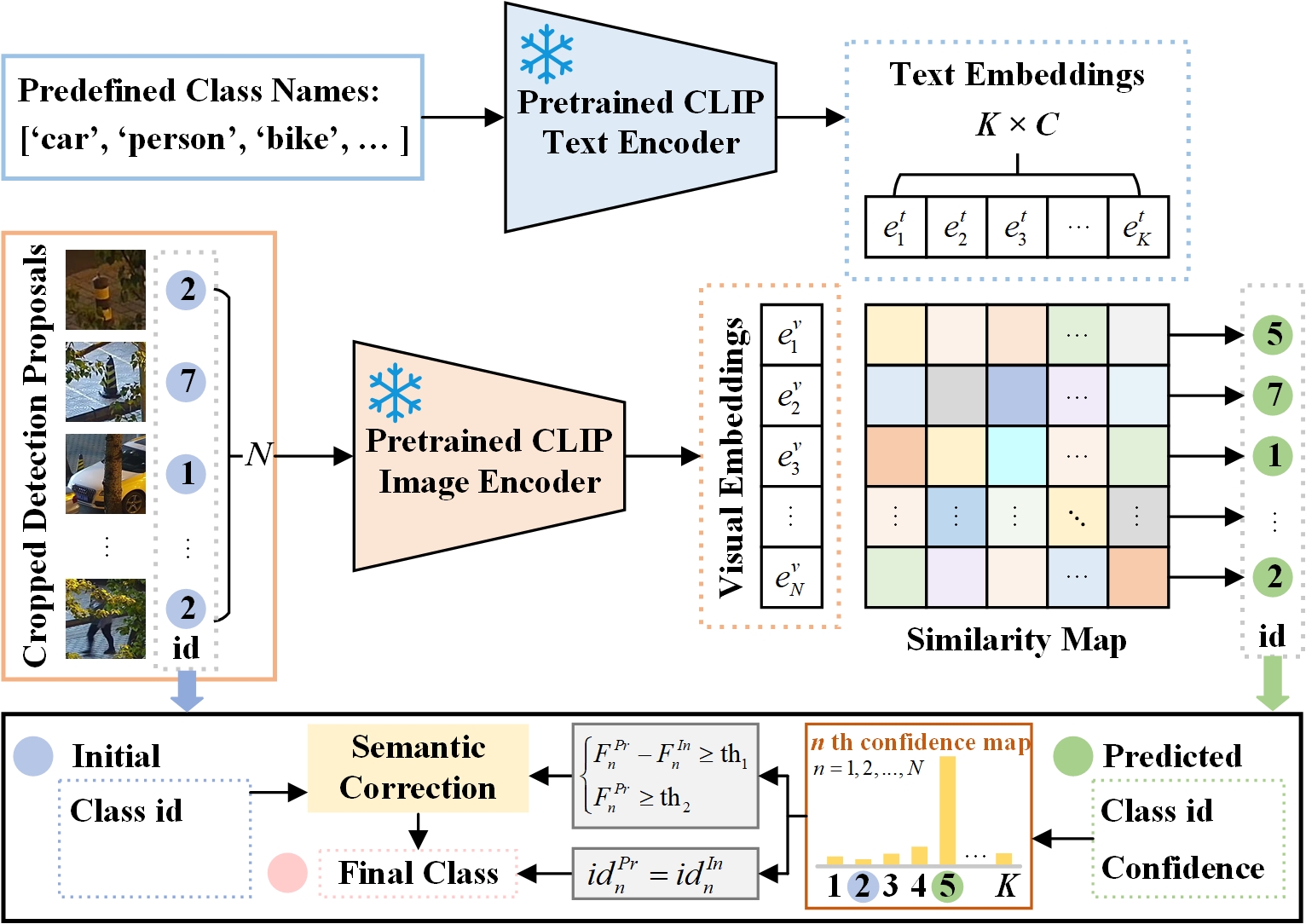}
  \caption{Semantic Consistency Correction Module.}
  \label{fig3}
\end{figure}

Due to potential detection errors for Grounding DINO in challenging scenes, we employ CLIP \cite{clip} model to rectify the semantic information by calculating the similarity of the initial detection proposals and all categories, seen in Fig. \ref{fig3}.

After detection with multiple prompts, this process yields $N$ detection proposals, which can be denoted as $v_1, v_2, \dots, v_N$, each corresponding to an initial class id $id_1^{In}, id_2^{In}, \dots, id_N^{In}$, respectively. 

Then, we leverage a pre-trained CLIP model to independently encode both the semantic categories and the image proposals, obtaining the corresponding visual embeddings $e_1^{v}, e_2^{v}, \dots, e_N^{v}$, which are collectively denoted as $e_n^{v}$ for $n=1, 2, \dots, N$, as well as the text embeddings $e_1^{t}, e_2^{t}, \dots, e_K^{t}$, denoted as $e_k^{t}$ for $k=1, 2, \dots, K$. The similarity scores between the visual and text embeddings are computed as follows:
\begin{equation}
F_{nk}=\frac{\exp \left \langle e_n^{v}, e_k^{t}\right \rangle}{1+\displaystyle\sum_{k=1}^{K}\exp \left \langle e_n^{v}, e_k^{t}\right \rangle}
\label{eq5}
\end{equation}
where $e_n^{v}$ represents the $n$-th visual embedding, $e_k^{t}$ denotes the $k$-th text embedding, and $F_{nk}$ corresponds to the predicted confidence of the $n$-th visual proposal belonging to the $k$-th class. The $\left \langle \cdot, \cdot \right \rangle$ represents the dot product operation. 

For each detection proposal, we then select the class with the highest prediction score $F_n^{Pr}=\max(F_{nk})$ as the predicted class label $id_n^{Pr}=k.\mathrm{index}(F_n^{Pr})$. If the predicted category $id_n^{Pr}$ matches the initial detection category $id_n^{In}$, it indicates that the semantic understanding is consistent, and no further correction is required. 

However, if the predicted category differs from the initial detection, we need to perform additional checks based on the following conditions. Firstly, we obtain the confidence score corresponding to the initial detection category:
\begin{equation}
F_n^{In}=F_{nk}\big|_{k=id_n^{In}}
\label{eq6}
\end{equation}

Then, we apply the following judgment criteria:

\begin{equation}
\left\{\begin{matrix}
F_n^{Pr}-F_n^{In}\geq \mathrm{th}_1 \\ 
F_n^{Pr}\geq \mathrm{th}_2
\end{matrix}\right.
\label{eq7}
\end{equation}
where $\mathrm{th}_1$ and $\mathrm{th}_2$ are two constant thresholds used to determine whether the predicted class should be updated. 

Overall, by incorporating this module, we can potentially improve the overall accuracy and robustness of the object classification task for the RGB-T images.

\subsection{Zero-shot Semantic Segmentation}
With the detection bounding boxes corresponding to each detection proposal obtained above, Open-RGBT can be extended to perform a variety of perception tasks, including zero-shot semantic segmentation. By integrating the Tokenize Anything via Prompting (TAP) model \cite{tap}, the model is able to segment the primary objects within each bounding box and generate descriptive captions for them. This process yields a set of masks $\left \{ \mathbf{M}_n \right \}_{n=1, 2, \dots, N}$ and a set of captions $\left \{ \mathbf{C}_n \right \}_{n=1, 2, \dots, N}$.  


\begin{figure}[!ht]
  \centering
  \setlength{\abovecaptionskip}{-0.2em}
  \includegraphics[width=0.8\linewidth]{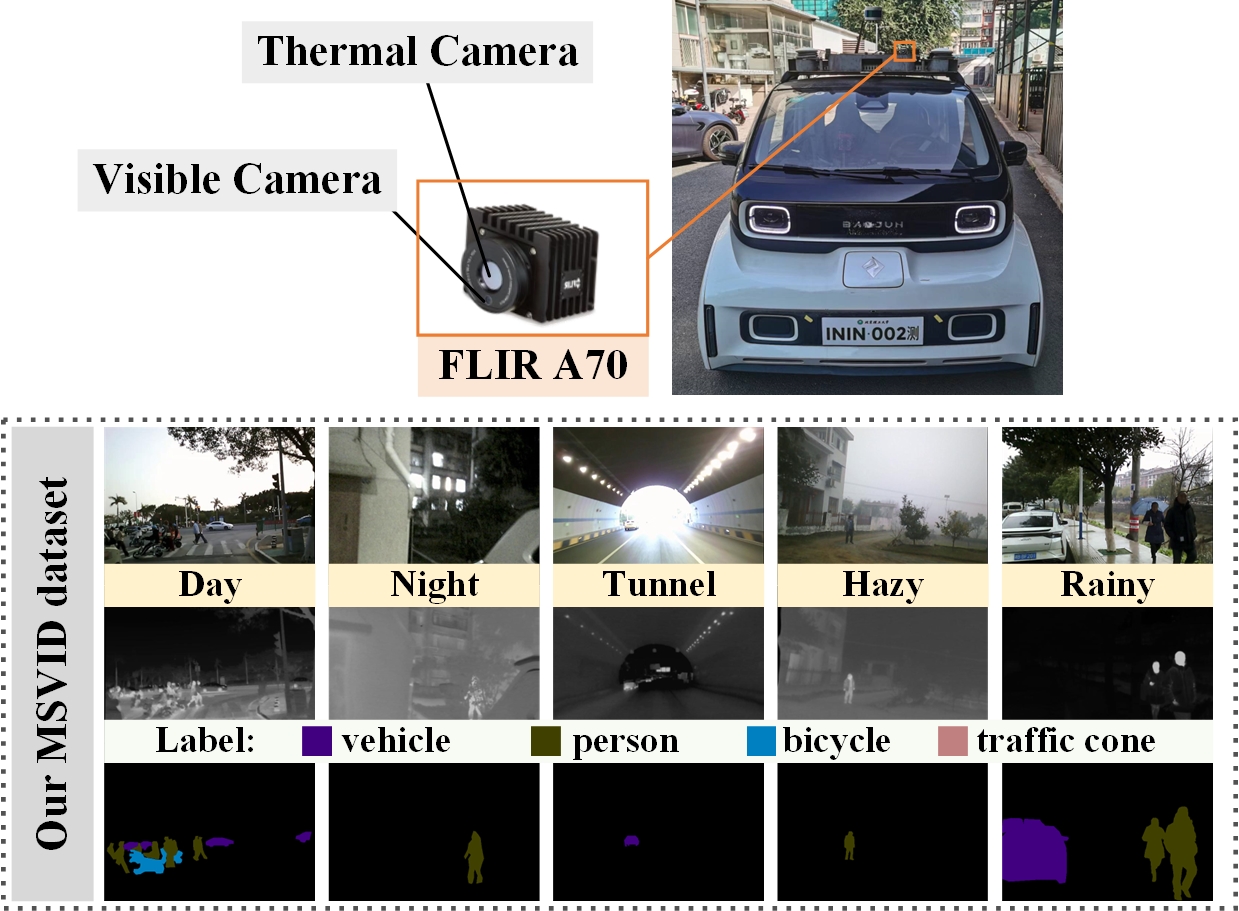}
  \caption{Our experimental platform and constructed MSVID dataset.}
  \vspace{-1.3em}
  \label{fig4}
\end{figure}

\section{EXPERIMENT}

In this section, we aim to validate Open-RGBT, through the following specific questions:

1) Without training any model, can Open-RGBT achieve RGB-T zero-shot semantic segmentation?

2) How well does Open-RGBT adapt to adverse weather and illumination-varying conditions?

3) What other potential scenes can be applied by this RGB-T open-vocabulary learning settings?

\subsection{RGB-T Zero-shot Semantic Segmentation}

\subsubsection{Datasets and Evaluation Metrics}
We adopt four commonly uses datasets for validation: MFNet \cite{mfnet}, PST900 \cite{pst900}, M3FD \cite{m3fd}, and Roadscene \cite{roadscene}. For the first two datasets, corresponding semantic labels are provided, while for the latter two, we annotate three familiar road participants (car, person, and bike). 

Furthermore, we use our own experimental vehicle platform to collect RGB-T image pairs under diverse environmental conditions, and construct our MSVID dataset with ground-truth semantic segmentation labels, seen in Fig. \ref{fig4}. For the evaluation metrics, we use the mean accuracy (mAcc) and the mean IoU (mIoU).

\begin{table*}
\caption{Comparison of qualitative semantic segmentation results (\%) on five diverse datasets.}
\renewcommand{\arraystretch}{0.9}
\begin{tabular}{p{1.2cm}|c|cccc|cccccc|p{0.5cm}p{0.5cm}} 
\hline
\multicolumn{2}{c|}{\multirow{3}{*}{\textbf{Method}}} & \multicolumn{4}{c|}{\textbf{Trained}} & \multicolumn{6}{c|}{\textbf{Zero-shot Testing}} & \multicolumn{2}{c}{\multirow{2}{*}{Mean}}  \\
\cline{3-12}
\multicolumn{2}{c|}{} & \multicolumn{2}{c}{MFNet} & \multicolumn{2}{c|}{PST900} & \multicolumn{2}{c}{M3FD} & \multicolumn{2}{c}{Roadscene} & \multicolumn{2}{c|}{MSVID} & \multicolumn{2}{c}{} \\
\multicolumn{2}{c|}{}& mAcc & mIoU & mAcc & mIoU & mAcc & mIoU & mAcc & mIoU & mAcc & mIoU & mAcc & mIoU  \\ 
\hline\hline
\multirow{9}{*}{\centering \textit{\textbf{Closed-set}}}
&MFNet \cite{mfnet}  & 45.10 & 39.70 & - & 57.02 & 56.26 & 33.42 & 35.86 & 30.48 & 39.67 & 27.10 & - & 37.54 \\
&RTFNet \cite{rtfnet} & 63.10 & 53.20 & 65.69 & 60.46 & 66.25 & 44.36 & 69.57 & 58.62 & 47.20 & 33.19 & 62.36 & 49.97 \\
&GMNet \cite{gmnet} & 74.10 & 57.30 & 89.61 & 84.12 & 65.93 & 47.22 & 71.46 & 53.36 & 53.64 & 40.66 & 70.95 & 56.63 \\
&EGFNet \cite{egfnet} & 72.70 & 54.80 & 94.02 & 78.51 & 68.76 & 51.33 & 82.97 & 65.86 & 67.84 & 45.41 & 75.36 & 59.18 \\
&CCFFNet \cite{ccffnet} & 68.30 & 57.20 & 86.90 & 82.10 & - & - & - & - & - & - & - & - \\
&LASNet \cite{LASnet} & 75.40 & 54.90 & 91.63 & 84.40 & 64.47 & 47.54 & 77.57 & 56.7 & 69.78 & 44.02 & 75.77 & 57.51 \\
&semanticRT \cite{semanticrt} & -  & 58.00 & 89.67 & 84.47 & 61.59 & 46.99 & 65.44 & 60.63 & 52.57 & 44.20 & - & 58.86 \\
&EAEFNet \cite{eaefnet} & \textbf{75.10} & 58.90 & 91.10 & 85.40 & 68.75 & 59.35 & 74.71 & 64.84 & 54.06 & 41.82 & 72.74 & 62.06 \\
&CAINet \cite{cainet} & 73.20 & 58.60 & \textbf{94.27} & 84.74 & 70.64 & 60.58 & 56.73 & 54.46 & 60.32 & 53.39 & 71.03 & 62.35 \\ 
\hline
\multirow{8}{*}{\parbox{1cm}{\centering \textit{\textbf{Open-vocabulary}}}}
& & \multicolumn{10}{c|}{\textbf{Zero-shot Testing}} \\
\cline{3-12}
&OV-seg \cite{ovseg} (RGB) & 41.45 & 21.92 & 73.82 & 32.13 & 46.46 & 32.06 & 69.95 & 34.32 & 60.23 & 18.23 & 58.38 & 27.73 \\
&OV-seg \cite{ovseg} (T) & 37.54 & 16.38 & 28.72 & 7.24 & 51.45 & 28.95 & 64.72 & 30.14 & 38.54 & 10.11 & 44.19 & 18.56 \\
&SEEM \cite{seem} (RGB) & 27.79 & 22.06 & 49.87 & 44.67 & 54.62 & 52.35 & 71.49 & 68.82 & 56.03 & 37.60 & 51.67 & 44.82 \\
&SEEM \cite{seem} (T) & 27.90 & 20.26 & 23.72 & 22.74 & 62.32 & 53.42 & 65.77 & 61.43 & 44.64 & 39.18 & 44.36 & 38.89 \\
&Grounded-SAM \cite{groundsam} (RGB) & 37.13 & 29.70 & 33.26 & 27.10 & 44.27 & 38.91 & 63.14 & 56.23 & 48.55 & 38.68 & 45.27 & 38.12 \\
&Grounded-SAM \cite{groundsam} (T) & 36.49 & 24.94 & 32.90 & 23.99 & 50.65 & 44.50 & 63.08 & 56.78 & 41.48 & 30.37 & 44.92 & 36.12 \\
&Ours (RGB-T) & 73.08 & \textbf{61.64} & 93.75 & \textbf{89.76} & \textbf{72.60} & \textbf{69.25} & \textbf{79.36} & \textbf{75.19} & \textbf{71.56} & \textbf{65.25} & \textbf{78.07} & \textbf{72.22} \\
\hline
\end{tabular}
\label{tab1}
\end{table*}

\subsubsection{Comparison on Multiple Datasets}
We compare our method with 9 closed-set RGB-T semantic segmentation methods (MFNet \cite{mfnet}, RTFNet \cite{rtfnet}, GMNet \cite{gmnet}, EGFNet \cite{egfnet}, CCFFNet \cite{ccffnet}, LASNet \cite{LASnet},  semanticRT \cite{semanticrt}, EAEFNet \cite{eaefnet}, and CAINet \cite{cainet}) and three open-vocabulary semantic segmentation methods (Ov-seg \cite{ovseg}, Seem \cite{seem}, and Grounded-SAM \cite{groundsam}). Given that these open-vocabulary models are single-modal, they are tested separately using both RGB and thermal images.
\begin{table}
\caption{Quantitative comparisons on mIoU (\%) of open-vocabulary methods on four categories of our MSVID dataset.}
\centering
\renewcommand{\arraystretch}{1}
\begin{tabular}{c|cccc} 
\hline
\textbf{Method} & \textbf{vehicle}& \textbf{person}& \textbf{bicycle}& \textbf{cone} \\ 
\hline\hline
OV-seg \cite{ovseg} (RGB)  & 17.85  & 1.76  & 2.00  & 0.02   \\
OV-seg \cite{ovseg} (T)    & 3.99   & 2.89  & 0.04  & 0.00   \\
SEEM \cite{seem} (RGB)     & 72.38  & 21.73 & 35.61 & 0.01   \\
SEEM \cite{seem} (T)       & 38.89  & 58.17 & 0.89  & 0.00   \\
Grounded-SAM \cite{groundsam} (RGB) & 30.92 & 45.01 & 20.57 & 0.00   \\
Grounded-SAM  \cite{groundsam} (T)  & 11.36 & 46.05 & 1.80  & 0.00   \\
Ours (RGB-T)& \textbf{78.50}& \textbf{69.30}& \textbf{46.71}& \textbf{35.81}\\
\hline
\end{tabular}
\label{tab2}
\end{table}

\textbf{Quantitative Comparisons.} Seen in Table \ref{tab1}, our method achieves the highest mAcc and mIoU across all methods, outperforming the second-best method by 2.30\% and 9.87\%, respectively. Compared to the closed-set approaches, though our model's mAcc is not optimal on the MFNet and PST900 datasets, the differences are marginal at 2.02\% and 0.52\%, respectively. Furthermore, to assess the zero-shot generalization capability, closed-set methods trained on the MFNet dataset are evaluated on other datasets, revealing a general decline in performance. In contrast, our integrated open-vocabulary detection model demonstrates superior generalization. Notably, compared to other open-vocabulary methods, our method demonstrates leading performance in various scenes, attributable in part to the integration of thermal modality.

\begin{table}
\caption{Zero-shot quantitative comparisons (\%) across adverse weathers and light-varying conditions on M3FD dataset.}
\centering
\renewcommand{\arraystretch}{0.9}
\begin{tabular}{c|p{0.5cm}p{0.5cm}p{0.5cm}p{0.5cm}p{0.9cm}|p{0.5cm}}
\hline
\textbf{Method} & \textbf{Day} & \textbf{Night} & \textbf{Rainy} & \textbf{Hazy} & \textbf{Exposure} & Mean \\
\hline\hline
\multicolumn{7}{c}{\textit{\textbf{Closed-set RGB-T Semantic Segmentation Methods}}} \\
\hline
MFNet \cite{mfnet} & 35.68 & 34.84 & 31.96 & 34.43 & 30.18 & 33.42 \\
RTFNet \cite{rtfnet} & 39.50 & 48.12 & 59.08 & 40.70 & 34.40 & 44.36 \\
GMNet \cite{gmnet} & 46.92 & 56.69 & 57.69 & 41.82 & 32.98 & 47.22 \\
EGFNet \cite{egfnet} & 57.07 & 59.14 & 59.86 & 39.92 & 40.66 & 51.33 \\
LASNet \cite{LASnet} & 53.55 & 55.26 & 57.29 & 31.53 & 40.05 & 47.54 \\
semanticRT \cite{semanticrt} & 43.04 & 57.81 & 54.36 & 37.46 & 42.29 & 46.99 \\
EAEFNet \cite{eaefnet} & 49.44 & 61.62 & 59.46 & 85.46 & 40.78 & 59.35 \\
CAINet \cite{cainet} & 56.26 & \textbf{68.64} & 69.85 & 51.76 & 56.40 & 60.58 \\
\hline
\multicolumn{7}{c}{\textit{\textbf{Open-vocabulary Semantic Segmentation Methods}}} \\
\hline
OV-seg \cite{ovseg} (RGB) & 44.32 & 36.93 & 42.55 & 13.74 & 22.76 & 32.06 \\
OV-seg \cite{ovseg} (T) & 32.28 & 41.55 & 34.09 & 21.14 & 15.69 & 28.95 \\
SEEM \cite{seem} (RGB) & 44.04 & 53.31 & 71.39 & 51.45 & 34.45 & 50.93  \\
SEEM \cite{seem} (T) & 40.65 & 57.80 & 66.09 & 59.05 & 30.73 & 50.86 \\
\begin{tabular}[c]{@{}c@{}}Grounded-SAM\\ \cite{groundsam} (RGB)\end{tabular} & 35.27 & 35.28 & 38.58 & 51.82 & 33.60 & 38.91 \\
\begin{tabular}[c]{@{}c@{}}Grounded-SAM\\ \cite{groundsam} (T)\end{tabular} & 38.19 & 37.59 & 40.76 & 79.79 & 26.17 & 44.50 \\
Ours (RGB-T) & \textbf{58.17} & 61.92 & \textbf{77.10} & \textbf{88.10} & \textbf{60.96}  & \textbf{69.25} \\
\hline
\end{tabular}
\label{tab3}
\end{table}

\begin{figure*}[!ht]
  \centering
  \setlength{\abovecaptionskip}{-0.5em}
  \includegraphics[width=\linewidth]{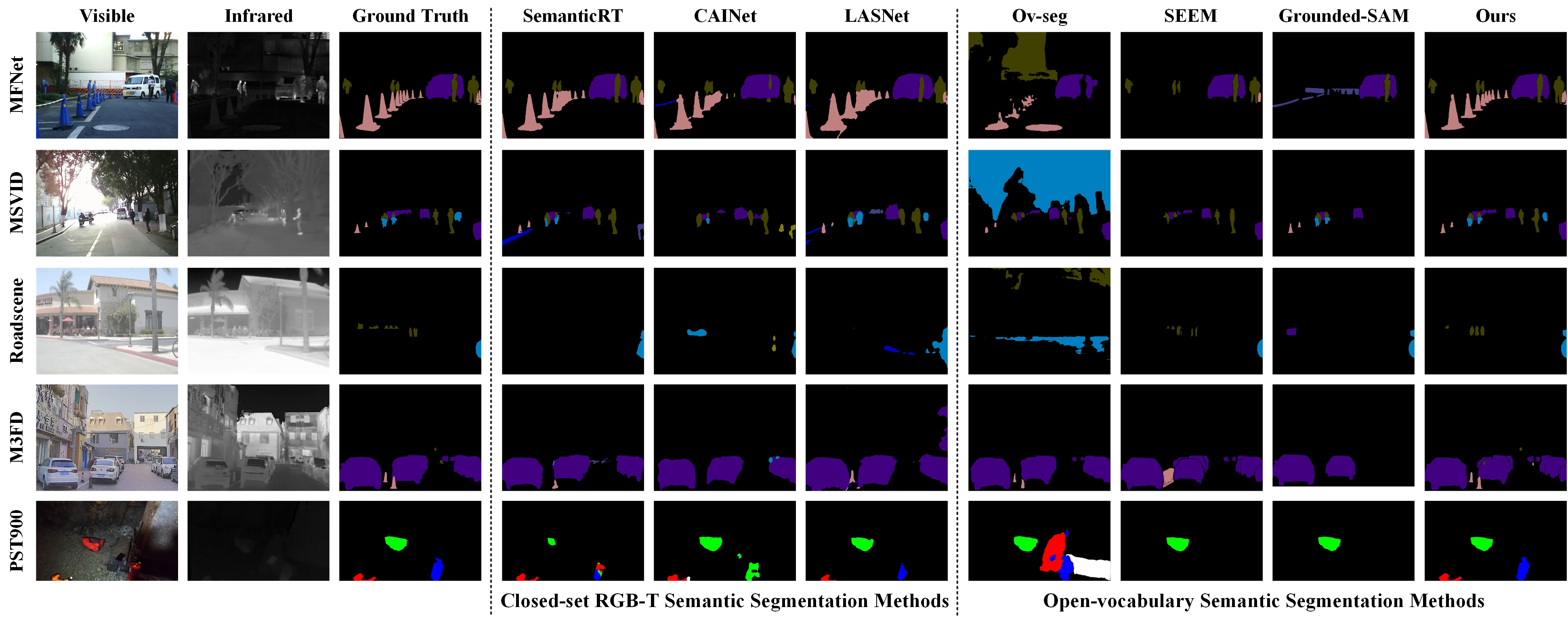}
  \caption{The sample qualitative results on multiple datasets. Please zoom in for best view.}
  \vspace{-1.5em}
  \label{fig5}
\end{figure*}

\textbf{Qualitative Comparisons.} Fig. \ref{fig5} displays several qualitative semantic segmentation results across diverse datasets. Our method demonstrates several notable advantages. Firstly, it can effectively delineate individual instances, even within cluttered scenes. As shown in the first row, where multiple color cones are present, competing methods often aggregate these into the entire area, but our approach distinctly separates each cone, owing to the precision of our detections. Secondly, our method excels in distinguishing objects from their background, as exemplified by the accurate delineation of the fire-extinguisher in the last row. Lastly, compared with Grounded-SAM, it is evident that there are numerous incorrect and missed detections, which states the efficacy of the incorporated modules in our model.

\subsubsection{Comparison of Specific Categories}
Table \ref{tab2} lists the quantitative comparisons of open-vocabulary methods in four specific categories of our MSVID dataset, where our method significantly outperforms others. In contrast, the other three methods perform better segmentation utilizing RGB modality than the thermal modality on vehicle and bicycle, while the opposite effect for person. Meanwhile, these methods fail to segment traffic cones. On the contrary, our method achieves the highest value due to the accurate detection of objects.

\subsection{Robustness to Adverse Open Environments}
To evaluate the robustness of each method, we conduct zero-shot tests across daytime and four challenging scenarios: night, rainy, hazy, and exposure, using the M3FD dataset. The target categories included vehicle, pedestrian, and bicycle. Quantitative comparisons based on the mIoU metric are presented in Table \ref{tab3}. Our method outperforms other methods in most cases, except for the night scene, where its performance is suboptimal compared to CAINet. This discrepancy arises because certain vehicles are obscured by elements such as pillars and trees. While our method effectively distinguishes objects from unrelated backgrounds, other methods may incorrectly classify these obscured areas as objects. The ground truth annotations, which depict complete vehicles, contribute to a decline in our method's accuracy. Furthermore, under night and hazy conditions, open-vocabulary methods utilizing the thermal modality exhibit better performance than those using the RGB modality, highlighting the importance of employing multiple modalities in open environments.

\begin{figure}[!ht]
  \centering
  \setlength{\abovecaptionskip}{-0.3em}
  \includegraphics[width=\linewidth]{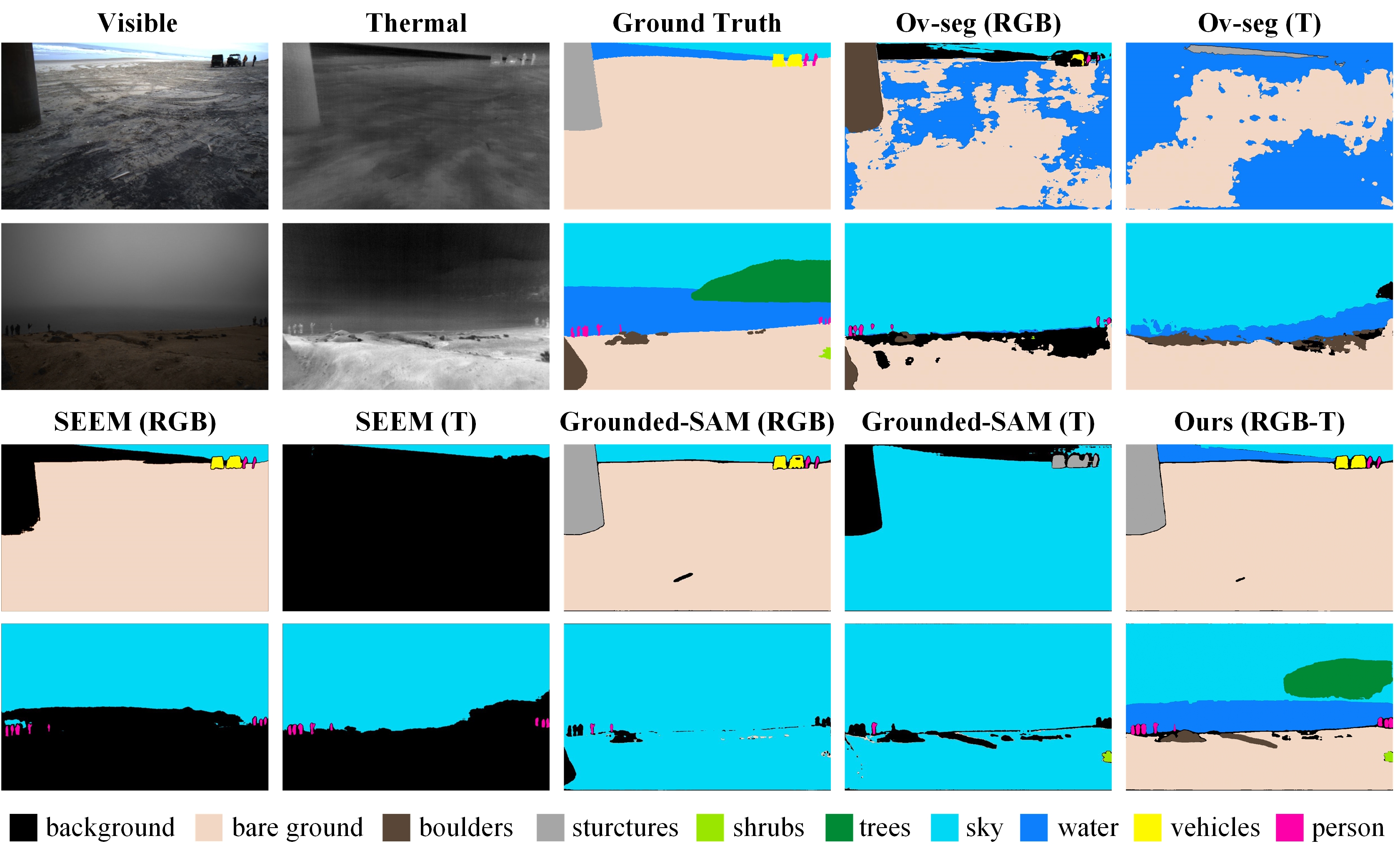}
  \caption{The sample qualitative results on wild dataset.}
  \vspace{-1em}
  \label{fig6}
\end{figure}


\subsection{Performance Analysis in the Wild}
We further explore the performance of open-vocabulary methods in wild scenarios, with the qualitative results depicted in Fig. \ref{fig6} based on the wild dataset \cite{cart}. It can be seen that our Open-RGBT can precisely segment the image. In contrast, Ov-seg struggles to effectively delineate these categories, resulting in incorrectly connected regions. SEEM fails to adequately separate objects from the background. Though Grounded-SAM demonstrates strong performance, it falls short in detecting small objects in the distance. Overall, our method exhibits robust performance among these approaches. 

\subsection{Ablation Study}
In this ablation study, we investigate the impact of the proposed method's components on MFNet and PST900 datasets, as summarized in Table \ref{tab5}. The baseline model utilizes RGB-T image pairs as input, employs Grounding DINO as the open-vocabulary object detector, and incorporates TAP as the segmentation model. 

\subsubsection{Effect of Visual Prompts} It is evident from the results that the inclusion of visual prompts leads to a significant improvement in the mAcc and mIoU metrics. Particularly for challenging classes, Grounding DINO may struggle to recognize objects and could potentially miss them. With the assistance of visual prompts, these challenging categories can be more reliably detected.

\subsubsection{Effect of SCCM} Similarly, the introduction of the Semantic Consistency Correction Module results in improvements in both the two metrics. This is because, in the presence of challenging classes, Grounding DINO might misinterpret the semantic meaning of the objects. With our module, the detection proposals can be refined, thereby enhancing the overall accuracy and robustness of the model.

\begin{table}
\caption{Ablation study of RGB-T open-vocabulary object detection strategies.}
\renewcommand{\arraystretch}{1}
\centering
\begin{tabular}{c|p{0.5cm}c|p{0.5cm}p{0.5cm}} 
\hline
\multirow{2}{*}{\textbf{Method}} & \multicolumn{2}{c|}{MFNet} & \multicolumn{2}{c}{PST900}  \\
& mAcc  & mIoU & mAcc  & mIoU  \\ 
\hline\hline
Grounding DINO &  55.80 & 38.30  & 78.10 & 58.00   \\ 
\hline
Grounding DINO + Visual Prompts & 69.33 & 57.18  & 90.62 & 84.57  \\
Grounding DINO + SCCM    & 60.51 & 45.26   & 82.20 & 69.14   \\ 
\hline
Ours & \textbf{73.08} & \textbf{61.64} & \textbf{93.75} & \textbf{89.76}  \\
\hline
\end{tabular}
\label{tab5}
\end{table}

\section{CONCLUSIONS}
In this paper, we present Open-RGBT, a multimodal open-vocabulary semantic segmentation model designed to address the limitations of traditional RGB-T segmentation models. By integrating open-vocabulary learning, our approach effectively generates detection proposals and utilizes visual prompts to enhance category understanding. The subsequent use of the CLIP model corrects semantic consistency, reducing ambiguity and improving the detection accuracy. Additionally, the inclusion of a unified foundation model for segmenting and captioning regions further enhances scene comprehension. Empirical evaluations confirm that Open-RGBT, operating as a zero-shot method, even exhibits superior performance compared to previous fully supervised methods, showcasing advantages in generalizing to challenging scenes. Future research will refine the model's capabilities and explore its applications in panoptic segmentation.


\bibliographystyle {unsrt} 
\bibliography{root}

\begin{thebibliography}{10}
\providecommand{\url}[1]{#1}
\csname url@samestyle\endcsname
\providecommand{\newblock}{\relax}
\providecommand{\bibinfo}[2]{#2}
\providecommand{\BIBentrySTDinterwordspacing}{\spaceskip=0pt\relax}
\providecommand{\BIBentryALTinterwordstretchfactor}{4}
\providecommand{\BIBentryALTinterwordspacing}{\spaceskip=\fontdimen2\font plus
\BIBentryALTinterwordstretchfactor\fontdimen3\font minus \fontdimen4\font\relax}
\providecommand{\BIBforeignlanguage}[2]{{%
\expandafter\ifx\csname l@#1\endcsname\relax
\typeout{** WARNING: IEEEtran.bst: No hyphenation pattern has been}%
\typeout{** loaded for the language `#1'. Using the pattern for}%
\typeout{** the default language instead.}%
\else
\language=\csname l@#1\endcsname
\fi
#2}}
\providecommand{\BIBdecl}{\relax}
\BIBdecl

\bibitem{book}
Y.~Yue and D.~Wang, \emph{Collaborative Perception, Localization and Mapping for Autonomous Systems}.\hskip 1em plus 0.5em minus 0.4em\relax Springer Nature, 2020, vol.~2.

\bibitem{robot}
S.~Ainetter and F.~Fraundorfer, ``End-to-end trainable deep neural network for robotic grasp detection and semantic segmentation from rgb,'' in \emph{2021 IEEE International Conference on Robotics and Automation (ICRA)}.\hskip 1em plus 0.5em minus 0.4em\relax IEEE, 2021, pp. 13\,452--13\,458.

\bibitem{remote}
J.~Jin, W.~Zhou, R.~Yang, L.~Ye, and L.~Yu, ``Edge detection guide network for semantic segmentation of remote-sensing images,'' \emph{IEEE Geoscience and Remote Sensing Letters}, vol.~20, pp. 1--5, 2023.

\bibitem{ovlearning}
J.~Wu, X.~Li, S.~Xu, H.~Yuan, H.~Ding, Y.~Yang, X.~Li, J.~Zhang, Y.~Tong, X.~Jiang \emph{et~al.}, ``Towards open vocabulary learning: A survey,'' \emph{IEEE Transactions on Pattern Analysis and Machine Intelligence}, 2024.

\bibitem{semanticrt}
W.~Ji, J.~Li, C.~Bian, Z.~Zhang, and L.~Cheng, ``Semanticrt: A large-scale dataset and method for robust semantic segmentation in multispectral images,'' in \emph{Proceedings of the 31st ACM International Conference on Multimedia}, 2023, pp. 3307--3316.

\bibitem{cainet}
Y.~Lv, Z.~Liu, and G.~Li, ``Context-aware interaction network for rgb-t semantic segmentation,'' \emph{IEEE Transactions on Multimedia}, 2024.

\bibitem{LASnet}
G.~Li, Y.~Wang, Z.~Liu, X.~Zhang, and D.~Zeng, ``Rgb-t semantic segmentation with location, activation, and sharpening,'' \emph{IEEE Transactions on Circuits and Systems for Video Technology}, vol.~33, no.~3, pp. 1223--1235, 2022.

\bibitem{llvip}
X.~Jia, C.~Zhu, M.~Li, W.~Tang, and W.~Zhou, ``Llvip: A visible-infrared paired dataset for low-light vision,'' in \emph{Proceedings of the IEEE/CVF international conference on computer vision}, 2021, pp. 3496--3504.

\bibitem{m3fd}
J.~Liu, X.~Fan, Z.~Huang, G.~Wu, R.~Liu, W.~Zhong, and Z.~Luo, ``Target-aware dual adversarial learning and a multi-scenario multi-modality benchmark to fuse infrared and visible for object detection,'' in \emph{Proceedings of the IEEE/CVF conference on computer vision and pattern recognition}, 2022, pp. 5802--5811.

\bibitem{sam}
A.~Kirillov, E.~Mintun, N.~Ravi, H.~Mao, C.~Rolland, L.~Gustafson, T.~Xiao, S.~Whitehead, A.~C. Berg, W.-Y. Lo \emph{et~al.}, ``Segment anything,'' in \emph{Proceedings of the IEEE/CVF International Conference on Computer Vision}, 2023, pp. 4015--4026.

\bibitem{sambad}
W.~Ji, J.~Li, Q.~Bi, T.~Liu, W.~Li, and L.~Cheng, ``Segment anything is not always perfect: An investigation of sam on different real-world applications,'' 2024.

\bibitem{mfnet}
Q.~Ha, K.~Watanabe, T.~Karasawa, Y.~Ushiku, and T.~Harada, ``Mfnet: Towards real-time semantic segmentation for autonomous vehicles with multi-spectral scenes,'' in \emph{2017 IEEE/RSJ International Conference on Intelligent Robots and Systems (IROS)}.\hskip 1em plus 0.5em minus 0.4em\relax IEEE, 2017, pp. 5108--5115.

\bibitem{roadscene}
H.~Xu, J.~Ma, Z.~Le, J.~Jiang, and X.~Guo, ``Fusiondn: A unified densely connected network for image fusion,'' in \emph{Proceedings of the AAAI conference on artificial intelligence}, vol.~34, no.~07, 2020, pp. 12\,484--12\,491.

\bibitem{rtfnet}
Y.~Sun, W.~Zuo, and M.~Liu, ``Rtfnet: Rgb-thermal fusion network for semantic segmentation of urban scenes,'' \emph{IEEE Robotics and Automation Letters}, vol.~4, no.~3, pp. 2576--2583, 2019.

\bibitem{pst900}
S.~S. Shivakumar, N.~Rodrigues, A.~Zhou, I.~D. Miller, V.~Kumar, and C.~J. Taylor, ``Pst900: Rgb-thermal calibration, dataset and segmentation network,'' in \emph{2020 IEEE international conference on robotics and automation (ICRA)}.\hskip 1em plus 0.5em minus 0.4em\relax IEEE, 2020, pp. 9441--9447.

\bibitem{gmnet}
W.~Zhou, J.~Liu, J.~Lei, L.~Yu, and J.-N. Hwang, ``Gmnet: Graded-feature multilabel-learning network for rgb-thermal urban scene semantic segmentation,'' \emph{IEEE Transactions on Image Processing}, vol.~30, pp. 7790--7802, 2021.

\bibitem{egfnet}
W.~Zhou, S.~Dong, C.~Xu, and Y.~Qian, ``Edge-aware guidance fusion network for rgb--thermal scene parsing,'' in \emph{Proceedings of the AAAI conference on artificial intelligence}, vol.~36, no.~3, 2022, pp. 3571--3579.

\bibitem{ccffnet}
W.~Wu, T.~Chu, and Q.~Liu, ``Complementarity-aware cross-modal feature fusion network for rgb-t semantic segmentation,'' \emph{Pattern Recognition}, vol. 131, p. 108881, 2022.

\bibitem{eaefnet}
M.~Liang, J.~Hu, C.~Bao, H.~Feng, F.~Deng, and T.~L. Lam, ``Explicit attention-enhanced fusion for rgb-thermal perception tasks,'' \emph{IEEE Robotics and Automation Letters}, 2023.

\bibitem{segnet}
V.~Badrinarayanan, A.~Kendall, and R.~Cipolla, ``Segnet: A deep convolutional encoder-decoder architecture for image segmentation,'' \emph{IEEE transactions on pattern analysis and machine intelligence}, vol.~39, no.~12, pp. 2481--2495, 2017.

\bibitem{deeplab}
L.-C. Chen, G.~Papandreou, I.~Kokkinos, K.~Murphy, and A.~L. Yuille, ``Deeplab: Semantic image segmentation with deep convolutional nets, atrous convolution, and fully connected crfs,'' \emph{IEEE transactions on pattern analysis and machine intelligence}, vol.~40, no.~4, pp. 834--848, 2017.

\bibitem{clip}
A.~Radford, J.~W. Kim, C.~Hallacy, A.~Ramesh, G.~Goh, S.~Agarwal, G.~Sastry, A.~Askell, P.~Mishkin, J.~Clark \emph{et~al.}, ``Learning transferable visual models from natural language supervision,'' in \emph{International conference on machine learning}.\hskip 1em plus 0.5em minus 0.4em\relax PMLR, 2021, pp. 8748--8763.

\bibitem{groundingdino}
S.~Liu, Z.~Zeng, T.~Ren, F.~Li, H.~Zhang, J.~Yang, C.~Li, J.~Yang, H.~Su, J.~Zhu \emph{et~al.}, ``Grounding dino: Marrying dino with grounded pre-training for open-set object detection,'' \emph{arXiv preprint arXiv:2303.05499}, 2023.

\bibitem{ovseg}
F.~Liang, B.~Wu, X.~Dai, K.~Li, Y.~Zhao, H.~Zhang, P.~Zhang, P.~Vajda, and D.~Marculescu, ``Open-vocabulary semantic segmentation with mask-adapted clip,'' in \emph{Proceedings of the IEEE/CVF Conference on Computer Vision and Pattern Recognition}, 2023, pp. 7061--7070.

\bibitem{groundsam}
T.~Ren, S.~Liu, A.~Zeng, J.~Lin, K.~Li, H.~Cao, J.~Chen, X.~Huang, Y.~Chen, F.~Yan \emph{et~al.}, ``Grounded sam: Assembling open-world models for diverse visual tasks,'' \emph{arXiv preprint arXiv:2401.14159}, 2024.

\bibitem{seem}
X.~Zou, J.~Yang, H.~Zhang, F.~Li, L.~Li, J.~Wang, L.~Wang, J.~Gao, and Y.~J. Lee, ``Segment everything everywhere all at once,'' \emph{Advances in Neural Information Processing Systems}, vol.~36, 2024.

\bibitem{lseg}
B.~Li, K.~Q. Weinberger, S.~Belongie, V.~Koltun, and R.~Ranftl, ``Language-driven semantic segmentation,'' \emph{arXiv preprint arXiv:2201.03546}, 2022.

\bibitem{align}
C.~Jia, Y.~Yang, Y.~Xia, Y.-T. Chen, Z.~Parekh, H.~Pham, Q.~Le, Y.-H. Sung, Z.~Li, and T.~Duerig, ``Scaling up visual and vision-language representation learning with noisy text supervision,'' in \emph{International conference on machine learning}.\hskip 1em plus 0.5em minus 0.4em\relax PMLR, 2021, pp. 4904--4916.

\bibitem{psf}
L.~Tang, H.~Zhang, H.~Xu, and J.~Ma, ``Rethinking the necessity of image fusion in high-level vision tasks: A practical infrared and visible image fusion network based on progressive semantic injection and scene fidelity,'' \emph{Information Fusion}, vol.~99, p. 101870, 2023.

\bibitem{trex}
Q.~Jiang, F.~Li, Z.~Zeng, T.~Ren, S.~Liu, and L.~Zhang, ``T-rex2: Towards generic object detection via text-visual prompt synergy,'' \emph{arXiv preprint arXiv:2403.14610}, 2024.

\bibitem{zs3net}
M.~Bucher, T.-H. Vu, M.~Cord, and P.~P{\'e}rez, ``Zero-shot semantic segmentation,'' \emph{Advances in Neural Information Processing Systems}, vol.~32, 2019.

\bibitem{zegformer}
J.~Ding, N.~Xue, G.-S. Xia, and D.~Dai, ``Decoupling zero-shot semantic segmentation,'' in \emph{Proceedings of the IEEE/CVF Conference on Computer Vision and Pattern Recognition}, 2022, pp. 11\,583--11\,592.

\bibitem{zsseg}
M.~Xu, Z.~Zhang, F.~Wei, Y.~Lin, Y.~Cao, H.~Hu, X.~Bai \emph{et~al.}, ``A simple baseline for zero-shot semantic segmentation with pre-trained vision-language model,'' \emph{arXiv preprint arXiv:2112.14757}, vol.~3, p.~2, 2021.

\bibitem{openseg}
G.~Ghiasi, X.~Gu, Y.~Cui, and T.-Y. Lin, ``Scaling open-vocabulary image segmentation with image-level labels,'' in \emph{European Conference on Computer Vision}.\hskip 1em plus 0.5em minus 0.4em\relax Springer, 2022, pp. 540--557.

\bibitem{rgbt234}
C.~Li, X.~Liang, Y.~Lu, N.~Zhao, and J.~Tang, ``Rgb-t object tracking: Benchmark and baseline,'' \emph{Pattern Recognition}, vol.~96, p. 106977, 2019.

\bibitem{lasher}
C.~Li, W.~Xue, Y.~Jia, Z.~Qu, B.~Luo, J.~Tang, and D.~Sun, ``Lasher: A large-scale high-diversity benchmark for rgbt tracking,'' \emph{IEEE Transactions on Image Processing}, vol.~31, pp. 392--404, 2021.

\bibitem{apfnet}
Y.~Xiao, M.~Yang, C.~Li, L.~Liu, and J.~Tang, ``Attribute-based progressive fusion network for rgbt tracking,'' in \emph{Proceedings of the AAAI Conference on Artificial Intelligence}, vol.~36, no.~3, 2022, pp. 2831--2838.

\bibitem{vipt}
J.~Zhu, S.~Lai, X.~Chen, D.~Wang, and H.~Lu, ``Visual prompt multi-modal tracking,'' in \emph{Proceedings of the IEEE/CVF conference on computer vision and pattern recognition}, 2023, pp. 9516--9526.

\bibitem{gmmt}
Z.~Tang, T.~Xu, X.~Wu, X.-F. Zhu, and J.~Kittler, ``Generative-based fusion mechanism for multi-modal tracking,'' in \emph{Proceedings of the AAAI Conference on Artificial Intelligence}, vol.~38, no.~6, 2024, pp. 5189--5197.

\bibitem{bat}
B.~Cao, J.~Guo, P.~Zhu, and Q.~Hu, ``Bi-directional adapter for multimodal tracking,'' in \emph{Proceedings of the AAAI Conference on Artificial Intelligence}, vol.~38, no.~2, 2024, pp. 927--935.

\bibitem{untrack}
Z.~Wu, J.~Zheng, X.~Ren, F.-A. Vasluianu, C.~Ma, D.~P. Paudel, L.~Van~Gool, and R.~Timofte, ``Single-model and any-modality for video object tracking,'' in \emph{Proceedings of the IEEE/CVF Conference on Computer Vision and Pattern Recognition}, 2024, pp. 19\,156--19\,166.

\bibitem{sdstrack}
X.~Hou, J.~Xing, Y.~Qian, Y.~Guo, S.~Xin, J.~Chen, K.~Tang, M.~Wang, Z.~Jiang, L.~Liu \emph{et~al.}, ``Sdstrack: Self-distillation symmetric adapter learning for multi-modal visual object tracking,'' in \emph{Proceedings of the IEEE/CVF Conference on Computer Vision and Pattern Recognition}, 2024, pp. 26\,551--26\,561.

\bibitem{cart}
C.~Lee, M.~Anderson, N.~Raganathan, X.~Zuo, K.~Do, G.~Gkioxari, and S.-J. Chung, ``Cart: Caltech aerial rgb-thermal dataset in the wild,'' \emph{arXiv preprint arXiv:2403.08997}, 2024.

\bibitem{tap}
T.~Pan, L.~Tang, X.~Wang, and S.~Shan, ``Tokenize anything via prompting,'' \emph{arXiv preprint arXiv:2312.09128}, 2023.

\bibitem{xunjiergbt}
X.~He, M.~Wang, T.~Liu, L.~Zhao, and Y.~Yue, ``Sfaf-ma: Spatial feature aggregation and fusion with modality adaptation for rgb-thermal semantic segmentation,'' \emph{IEEE Transactions on Instrumentation and Measurement}, vol.~72, pp. 1--10, 2023.

\bibitem{vifnet}
M.~Yu, T.~Cui, H.~Lu, and Y.~Yue, ``Vifnet: An end-to-end visible-infrared fusion network for image dehazing,'' \emph{Neurocomputing}, p. 128105, 2024.

\bibitem{yueicra}
Y.~Yue, C.~Yang, J.~Zhang, M.~Wen, Z.~Wu, H.~Zhang, and D.~Wang, ``Day and night collaborative dynamic mapping in unstructured environment based on multimodal sensors,'' in \emph{2020 IEEE international conference on robotics and automation (ICRA)}.\hskip 1em plus 0.5em minus 0.4em\relax IEEE, 2020, pp. 2981--2987.

\end{thebibliography}

\end{document}